\documentclass{article}

\usepackage[dblblindworkshop,final]{neurips_2026}
\workshoptitle{Foundations of Language Model Security}

\usepackage[utf8]{inputenc}
\usepackage[T1]{fontenc}

\usepackage[hypertexnames=false]{hyperref}
\usepackage{url}
\usepackage{booktabs}
\usepackage{amsfonts}
\usepackage{amsmath}
\usepackage{amssymb}
\usepackage{nicefrac}
\usepackage{microtype}
\usepackage{xcolor}
\usepackage{graphicx}
\usepackage{float}

\title{System-Prompt Anchoring with Cross-Attention Layers}

\author{%
  Li Lixing \\
  Cornell University \\
  Ithaca, NY 14853 \\
  \texttt{ll963@cornell.edu} \\
}

\begin{document}

\maketitle

\begin{abstract}
  Cross-attention provides a dedicated route from a selected information source into a model's computation, but the effect of where that route is inserted remains underexplored. We study this question when the source is a privileged system-prompt span. We insert Cross-Attention Layer (CAL) blocks between the system prompt and text while keeping the causal-decoder backbone frozen. A ten-configuration sweep on a 1.5B backbone shows that performance is task-dependent and strongly affected by placement: later placements are generally more effective and parameter-efficient.
  In an 8B transfer study, we train the selected \textsc{late8th} configuration and compare it with parameter-matched adaptation baselines. Across the evaluated benchmarks, the effects remain task-dependent: cross-attention changes instruction-following and security behavior while largely preserving general-task performance. Together, these experiments characterize placement as an important design variable when injecting system-prompt information through cross-attention.
\end{abstract}

\section{Introduction}
\label{sec:intro}

Modern LLM deployment relies on a \emph{system prompt} to establish behavioral constraints and safety rules, yet standard Transformers process all tokens through a uniform causal self-attention mechanism~\cite{vaswani2017attention} that treats privileged instructions and untrusted user data with equal priority---a structural mismatch we term, following \citet{wallace2025instruction}, the Instruction Hierarchy problem.

This uniform priority yields two critical problems. First, models suffer from prompt injection, where adversarial user inputs overwrite the system prompt's rules and successfully hijack the model's objective \cite{perez2022ignore, greshake2023indirect}. Second, models suffer from \emph{context drift} (the gradual erosion of system instructions over extended or multi-turn sequences), progressively losing adherence to the system prompt's constraints.

We study the effect of injecting the system prompt through cross-attention: does a dedicated route to the privileged span change performance across security and instruction-following tasks, and how does the placement of the injected blocks affect that performance? We call each inserted module a \emph{Cross-Attention Layer} (CAL) block. A CAL block projects queries from the full residual stream and keys and values from the system-prompt span, while leaving the backbone's ordinary self-attention and residual paths intact.

We train zero-initialized CAL blocks on a frozen Qwen2.5-1.5B-Instruct backbone using a 50,000-example general SFT corpus and sweep ten placement configurations. The results are task-dependent: later placements generally provide stronger instruction adherence and a favorable performance--parameter trade-off. We then transfer the balanced, parameter-efficient \textsc{late8th} configuration to an 8B backbone, where it shows clearer gains in instruction adherence and resistance to many-shot jailbreaks while remaining task-dependent on other security evaluations.

\noindent Our contributions are:
\begin{enumerate}
  \item \textbf{System-prompt cross-attention training}: we train CAL blocks that route a system-prompt span into a frozen causal decoder and evaluate the resulting model across general, instruction-following, and security tasks.
  \item \textbf{Placement study}: we compare ten CAL placement strategies on a 1.5B backbone, and show that placement effects are substantial but task-dependent.
  \item \textbf{8B transfer study}: we transfer the selected \textsc{late8th} configuration to an 8B backbone and compare it with parameter-matched adaptation baselines.
\end{enumerate}

\section{Related work}
\label{sec:related}

\paragraph{Cross-attention for information integration.}
Encoder--decoder Transformers condition generation on an encoded input through cross-attention~\citep{vaswani2017attention,raffel2019t5}; RETRO integrates retrieved text through chunked cross-attention~\citep{borgeaud2022retro}; and Flamingo~\citep{alayrac2022flamingo} and BLIP-2~\citep{li2023blip2} connect frozen language models to visual representations. ControlNet likewise conditions a frozen diffusion backbone through zero-initialized branches~\citep{zhang2023controlnet}. CAL blocks follow this general pattern but source keys and values from a privileged span within the same causal sequence. Our focus is not a new cross-attention architecture, but how injection depth affects the use of that route.

\paragraph{Prompt injection and behavioral alignment.}
Prompt injection~\citep{perez2022ignore,greshake2023indirect} exploits conflicts between trusted instructions and untrusted content. SecAlign~\citep{piet2024secalign} and StruQ~\citep{chen2025struq} address this behavior through preference optimization and structured queries; Instruction Hierarchy training~\citep{wallace2025instruction} explicitly trains models to prioritize privileged instructions. AIR injects hierarchy signals into intermediate representations~\citep{kariyappa2025air}, while ASIDE separates instructions and data at the embedding level~\citep{zverev2026aside}. CAL instead routes the system span through cross-attention and studies how the route's placement changes behavior; it does not block attacker-controlled content from the ordinary self-attention path.

\paragraph{Mechanistic interpretability of rule representations.}
Representation engineering~\citep{zou2023representation} and analyses of feed-forward computation~\citep{geva2022transformer} motivate examining how behavioral information changes across network depth. We therefore sweep CAL placement and measure the magnitude of the trained blocks' residual updates in Section~\ref{sec:magnitude}; these measurements characterize adapter use rather than locating rules in the untouched backbone.

\section{System-prompt routing with cross-attention}
\label{sec:arch}

CAL augments a frozen causal decoder with a dedicated route to the system prompt. We insert blocks after selected decoder layers; each block lets the full residual stream query the system-prompt span through causal cross-attention and then applies a SwiGLU feed-forward sublayer~\citep{shazeer2020swiglu}. Varying the insertion layers changes where this privileged route can influence the computation while leaving the backbone unchanged.

\begin{figure}[t]
  \centering
  \includegraphics[width=0.85\linewidth]{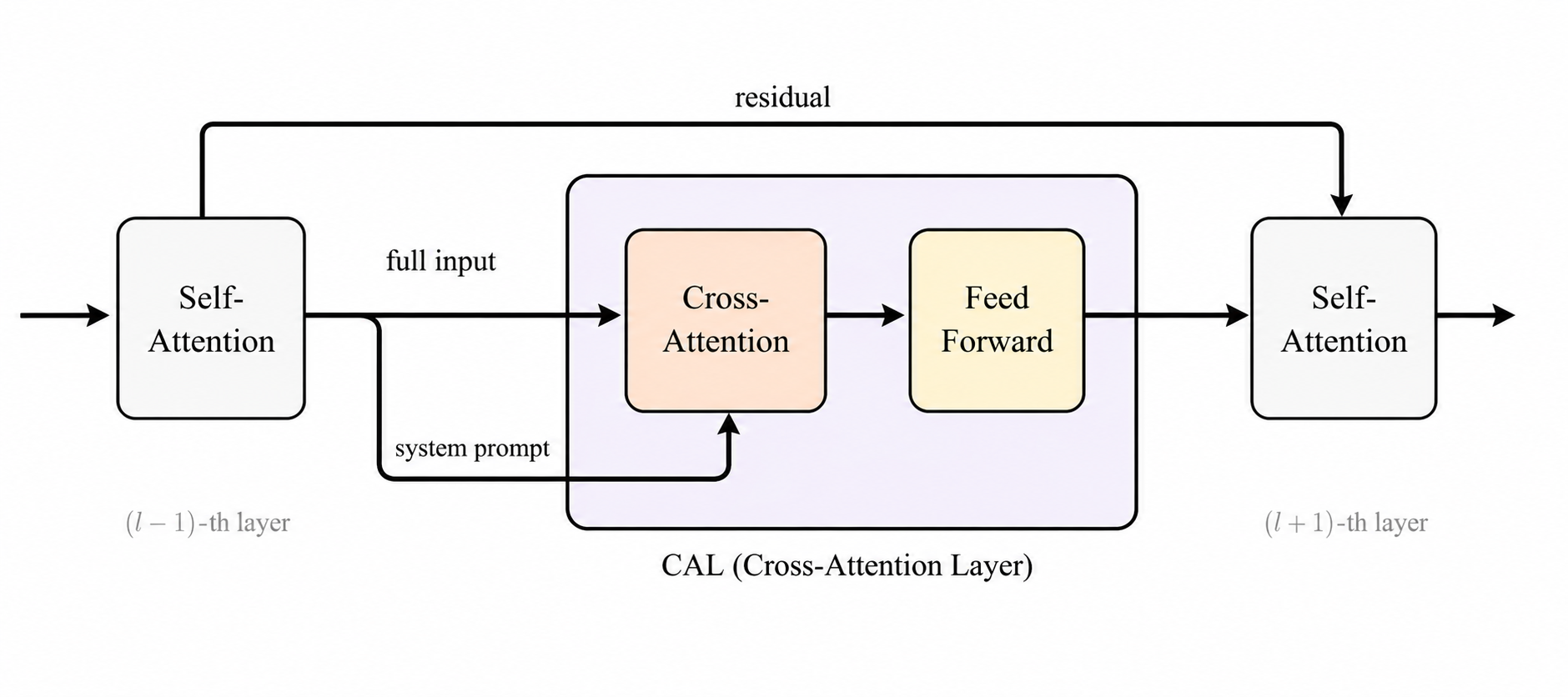}
  \caption{\textbf{System-prompt cross-attention.} A CAL block performs cross-attention between the full input and the system-prompt span. A feed-forward sublayer then processes the residual update before the next frozen decoder layer; the ordinary residual path remains available.}
  \label{fig:cal_architecture}
\end{figure}

\paragraph{Causal cross-attention.}
Let $\mathbf{X}$ be the residual stream entering a CAL block and $[s,e)$ the system-prompt span. After normalization, queries come from the full sequence, whereas keys and values come only from that span:
\[
  \mathbf{Q} = \bar{\mathbf{X}}\mathbf{W}_Q, \quad \mathbf{K} = \bar{\mathbf{X}}_{[s:e]}\mathbf{W}_K, \quad \mathbf{V} = \bar{\mathbf{X}}_{[s:e]}\mathbf{W}_V
\]
where $\bar{\mathbf{X}}=\mathrm{RMSNorm}(\mathbf{X})$~\citep{zhang2019rmsnorm}. A query at absolute position $i$ may attend to system-span slot $j$ only when $s+j\leq i$:

\[
  M_{i,j} = \begin{cases} 0 & \text{if } j \leq i - s \\ -\infty & \text{otherwise} \end{cases}
\]

Thus, tokens within the system prompt see only its causal prefix, while subsequent user-context and generated tokens can access the complete span without future-token leakage. The residual stream is updated as

\[
  \mathbf{X}' = \mathbf{X} + \text{Attention}(\mathbf{Q}, \mathbf{K}, \mathbf{V}, \mathbf{M})\mathbf{W}_O
\]

\paragraph{Feed forward.}
The attention update is followed by a normalized SwiGLU residual sublayer:
\[
  \mathbf{X}'' = \mathbf{X}' + \mathrm{SwiGLU}(\mathrm{RMSNorm}(\mathbf{X}'))\mathbf{W}_D
\]

\paragraph{Training.}
We freeze the backbone and train only the CAL parameters. Zero-initializing the output projections of both sublayers, $\mathbf{W}_O$ and $\mathbf{W}_D$, makes every inserted block an identity map at initialization. All configurations use the same 50,000-example general-purpose SFT corpus; Appendix~\ref{app:data} gives the dataset details.

\section{Ablation study: placement strategy on 1.5B backbone}
\label{sec:micro}

The primary objective of this section is to study how CAL placement affects performance on a 1.5B backbone. We test blocks at multiple network depths and compare their effects across general, instruction-following, and security tasks. We use the 28-layer \textbf{Qwen2.5-1.5B-Instruct}~\citep{qwen2025} as a frozen backbone to make the placement sweep computationally feasible.

\subsection{Experiment setup}
\label{sec:setup}

\paragraph{Control baselines.}
To separate the effect of the training corpus and added parameters from system-span routing, we evaluate against two baselines. Both are matched to the tunable parameter count of the selected \textsc{late8th} configuration and trained on the same dataset (the LoRA rank derivation is detailed in Appendix~\ref{app:lora}):
\begin{itemize}
  \item \textbf{LoRA Baseline}: A standard full-network LoRA. This baseline shows the effect of distributed adaptation to the same corpus, including narrow-corpus fine-tuning penalties.
  \item \textbf{ParallelMLP Baseline}: MLP adapters are injected parallel to the same self-attention blocks as \textsc{late8th}, following the parallel-adapter formulation of He et al.~\citep{he2021towards}. We zero-initialize them, match their parameter count, and train them on a frozen backbone. This baseline controls for trainable capacity, adapter location, and backbone freezing, but it is not a matched information-routing control because its computation differs from CAL.
\end{itemize}

\paragraph{Configurations.}
To understand how placement changes performance, we train ten CAL configurations across the 28-layer backbone. The full list is detailed in Appendix~\ref{app:config}.

\subsection{Empirical results and structural trade-offs}
\label{sec:results}

\begin{figure}[t]
  \centering
  \includegraphics[width=\textwidth]{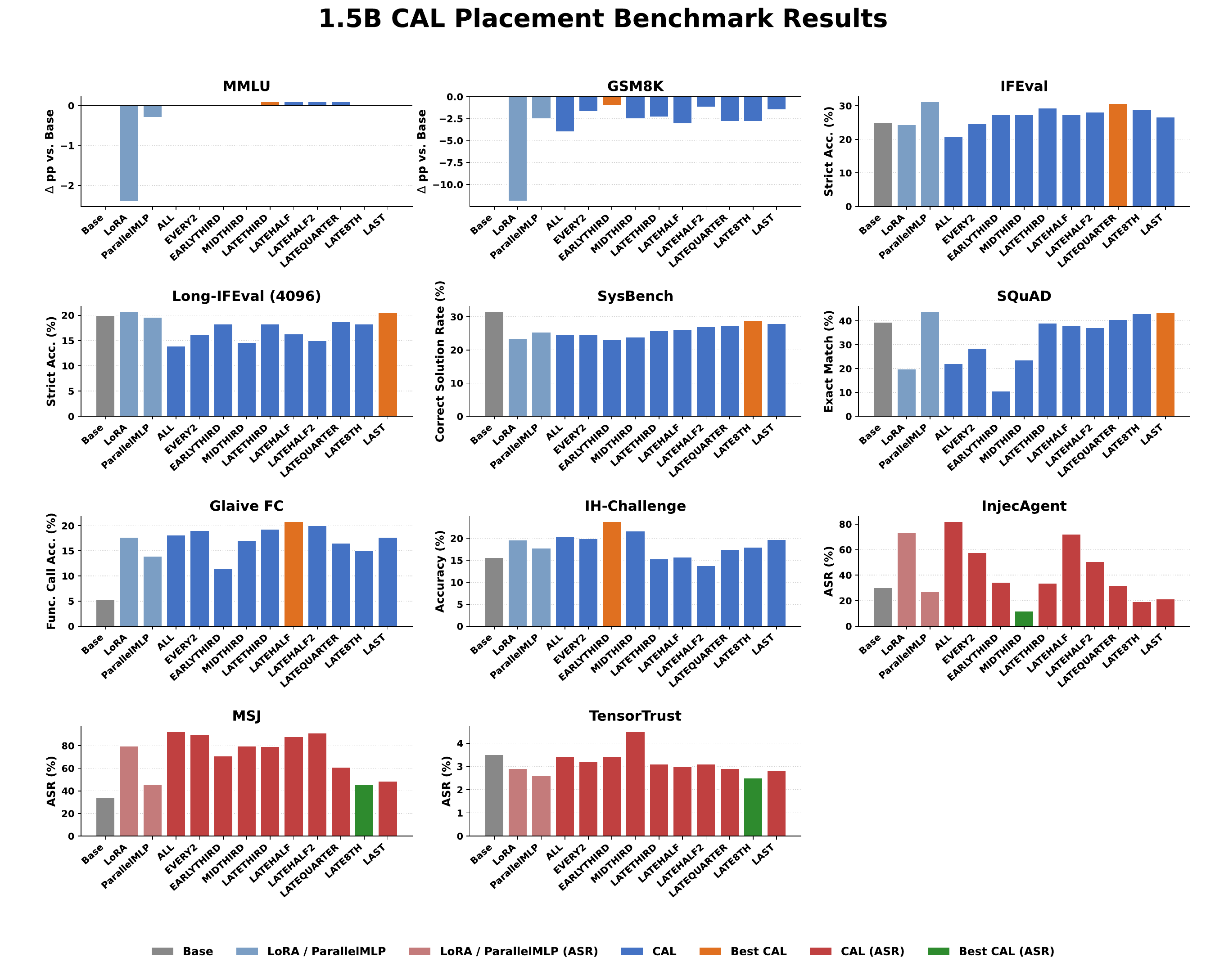}
  \caption{\textbf{1.5B CAL placement benchmark results.} Results across ten CAL placement configurations on Qwen2.5-1.5B-Instruct. MMLU and GSM8K are plotted as percentage-point deviations from the base model; other panels show absolute scores. Blue denotes task accuracy (higher is better), and red denotes attack success rate (ASR; lower is better).}
  \label{fig:barcharts}
\end{figure}

\paragraph{Evaluation.}
We measure general capability with MMLU~\citep{hendrycks2021mmlu} and GSM8K~\citep{cobbe2021gsm8k}; instruction following with IFEval~\citep{zhou2023ifeval}, SysBench~\citep{sysbench}, SQuAD~\citep{rajpurkar2018know}, and Glaive FC~\citep{glaive2023functioncalling}; and security with IH-Challenge~\citep{ihchallenge}, InjecAgent~\citep{zhan2024injecagent}, Many-Shot Jailbreaking (MSJ)~\citep{anil2024msj}, and TensorTrust~\citep{tensortrust}. Long-IFEval extends IFEval by placing its instruction in the system prompt and padding the user turn with essay text~\citep{kamradt2023needle,graham2026essays} to test strict accuracy at 4K--32K tokens; Appendix~\ref{app:longifeval} gives the construction details.

\paragraph{Placement effects.}
Figure~\ref{fig:barcharts} shows a clear but non-uniform depth effect. Sparse late placements generally offer the strongest performance--parameter trade-off, whereas early or dense insertion performs poorly on several instruction-following and security tasks. The exceptions are informative: different depths lead IFEval, Glaive FC, IH-Challenge, and InjecAgent, so no placement dominates every task. Within the late placements, distributing a small number of blocks is a stronger default than concentrating the intervention in one final block.

\paragraph{Matched controls.}
LoRA exposes the effect of adapting the full model to the narrow SFT corpus: it loses substantial MMLU and GSM8K performance and sharply increases attack success on InjecAgent and MSJ. The frozen-backbone adapters preserve general capability more closely. Relative to ParallelMLP, \textsc{late8th} is comparable on basic formatting and extraction, stronger on SysBench and Glaive FC, and narrowly better on all four 1.5B security evaluations. These differences support a benefit from system-span routing beyond parameter count and late-stage adaptation alone.

\begin{figure}[t]
  \centering
  \includegraphics[width=\textwidth]{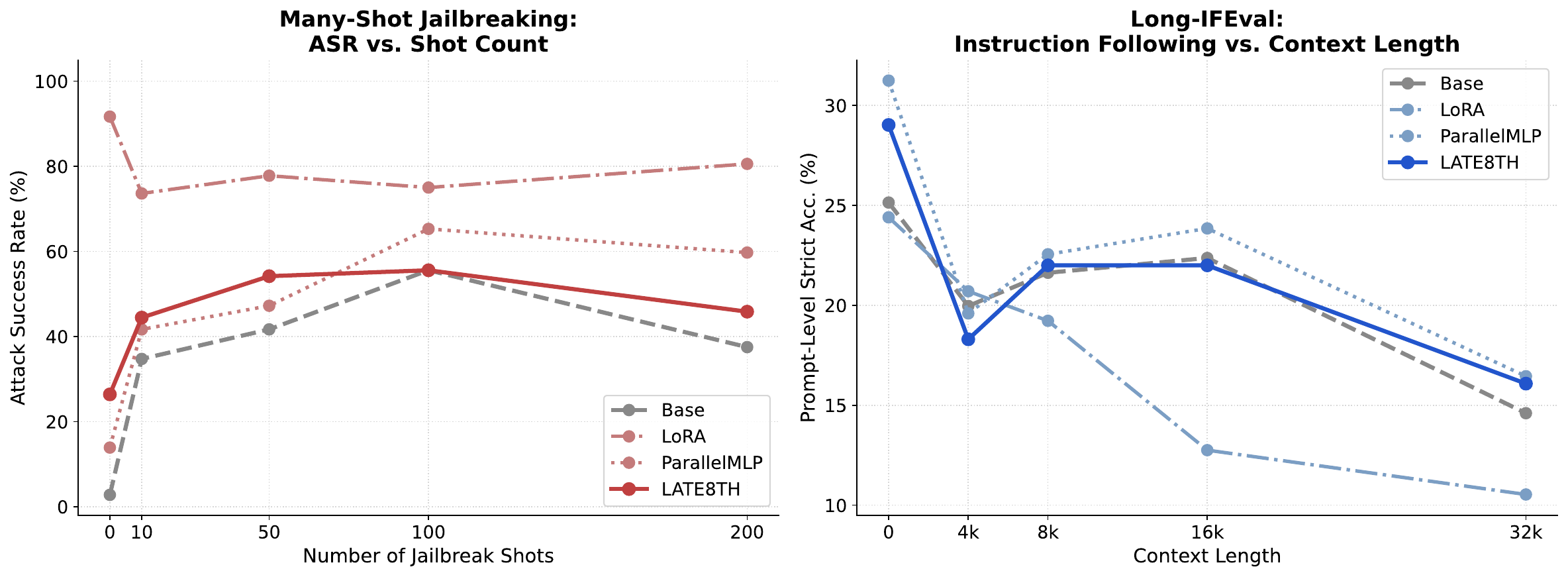}
  \caption{Structural resistance to adversarial and contextual degradation. \textbf{Left}: attack success rate (ASR) as a function of in-context many-shot jailbreak demonstrations (lower is better). \textbf{Right}: Long-IFEval strict accuracy over extended context windows (4k to 32k tokens).}
  \label{fig:degradation}
\end{figure}

\paragraph{Contextual degradation.}
Figure~\ref{fig:degradation} increases the in-context pressure through additional jailbreak demonstrations and longer user contexts. LoRA degrades rapidly outside the training distribution, which contains neither extreme-length nor adversarial multi-turn examples. At 200-shot MSJ, \textsc{late8th} outperforms ParallelMLP; on Long-IFEval, it degrades smoothly at extreme lengths but does not surpass ParallelMLP in absolute accuracy.

We select \textbf{\textsc{late8th}} as a balanced, parameter-efficient configuration for the 8B study. It leads the CAL configurations on MSJ and TensorTrust and remains competitive on instruction-following tasks, while other placements lead IH-Challenge, InjecAgent, IFEval, Long-IFEval, SQuAD, and Glaive FC.

\subsection{Mechanistic activation analysis}
\label{sec:magnitude}

Prior work shows that behavioral representations and feed-forward computation vary across network depth~\citep{zou2023representation,geva2022transformer}. To test whether CAL exhibits a corresponding depth-dependent utilization pattern, we measure each inserted block's residual update relative to its input:
$100\,\frac{\mathbb{E}_{b,t}\lVert\mathbf{X}''_{b,t}-\mathbf{X}_{b,t}\rVert_2}{\mathbb{E}_{b,t}\lVert\mathbf{X}_{b,t}\rVert_2}$,
where $\mathbf{X}$ is the input to the block, $\mathbf{X}''$ is its output, and the expectations average token-wise norms over batch and sequence positions. The measure includes both the cross-attention and SwiGLU updates and characterizes use of the inserted route rather than locating rules in the frozen backbone.

\begin{figure}[t]
  \centering
  \includegraphics[width=\textwidth]{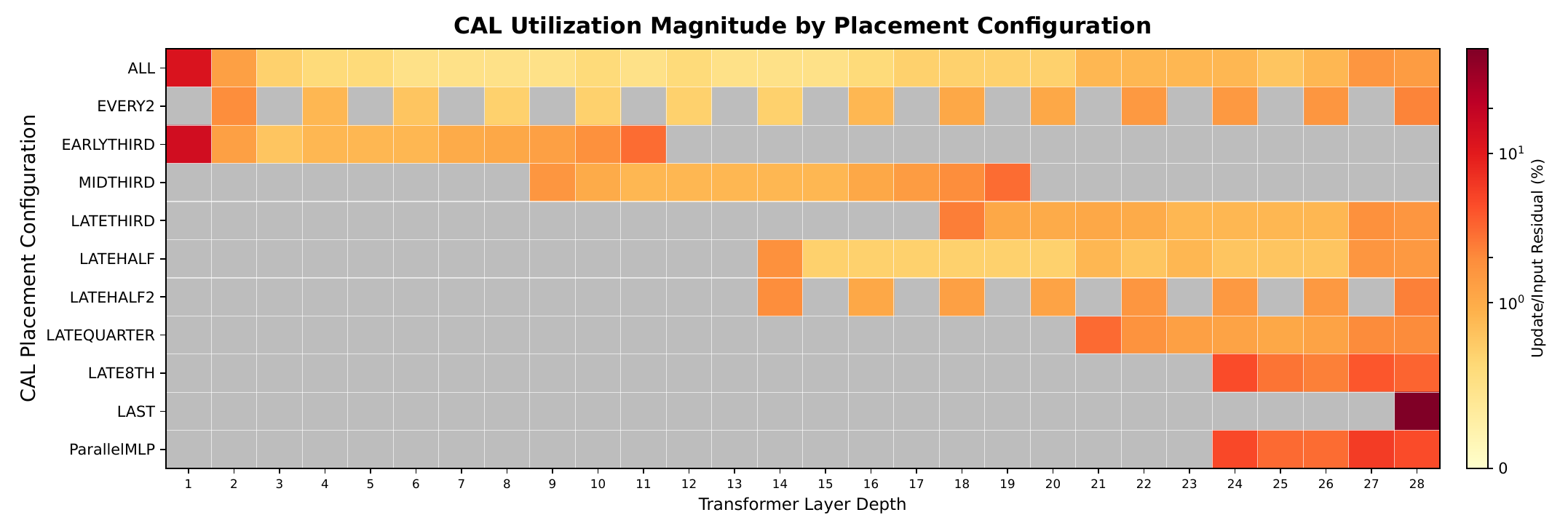}
  \caption{Mechanistic activation magnitude heatmap. Each cell shows the inserted-block update/input-residual $\ell_2$ norm ratio (\%) for one configuration and transformer depth; gray cells indicate no inserted block. Multi-block configurations exhibit a recurring within-window pattern, with larger updates near the beginning and end of the available window. ParallelMLP has a late-stage profile similar to \textsc{late8th}.}
  \label{fig:cal_magnitude}
\end{figure}

Figure~\ref{fig:cal_magnitude} reveals a recurring within-window pattern: update magnitude spikes at the first available block, falls through the middle, and rises again toward the end. The same shape appears in \textsc{earlythird} and \textsc{midthird} even though their windows occupy different absolute depths, while strong late configurations maintain updates above $1\%$ near the end of the network. This alignment across placements suggests that learned utilization depends on both absolute transformer depth and a block's relative position within the available injection window.

Magnitude is not itself a performance proxy. Dense early placements make large initial updates but perform poorly on several benchmarks, and the single-block \textsc{last} configuration produces a large update without being the strongest model. The activation pattern instead complements the behavioral results: effective configurations distribute moderate updates across late layers, consistent with depth-sensitive accounts of behavioral computation while showing how the added system-prompt route uses the locations made available to it.

\section{Transfer evaluation at 8B}
\label{sec:macro}

We test whether the \textsc{late8th} configuration selected on Qwen2.5-1.5B transfers to Llama-3.1-8B-Instruct. We compare it with two matched baselines: LoRA measures full-network adaptation to the same corpus, while ParallelMLP controls for parameter count, late-stage placement, and backbone freezing. The \textsc{late8th}--ParallelMLP comparison tests the contribution of the dedicated system-prompt route under otherwise matched conditions.

\begin{table}[h]
  \centering
  \caption{Benchmark results on 8B models. All values are absolute percentages (\%). $\uparrow$/$\downarrow$ indicates whether higher or lower scores are better. Bold figures denote the better score between ParallelMLP and CAL; $^{**}$ ($p<0.01$) and $^{*}$ ($p<0.05$) denote statistical significance under a two-sided proportion $z$-test.}
  \label{tab:8b-macro}
  \vspace{0.1cm}
  \resizebox{\textwidth}{!}{%
    \begin{tabular}{l cc ccccc cccc}
      \toprule
                             & \multicolumn{2}{c}{\textbf{General}} & \multicolumn{5}{c}{\textbf{Instruction-Following}} & \multicolumn{4}{c}{\textbf{Security}}                                                                                                                                                                                                                                                          \\
      \cmidrule(lr){2-3}\cmidrule(lr){4-8}\cmidrule(lr){9-12}
      \textbf{Model}         & \textbf{MMLU} $\uparrow$             & \textbf{GSM8K} $\uparrow$                          & \textbf{IFEval} $\uparrow$            & \textbf{L-IFEval} $\uparrow$ & \textbf{SysBench} $\uparrow$ & \textbf{SQuAD} $\uparrow$ & \textbf{Glaive} $\uparrow$ & \textbf{IH-Chal.} $\uparrow$ & \textbf{InjecAgent} $\downarrow$ & \textbf{MSJ} $\downarrow$ & \textbf{TensorTrust} $\downarrow$ \\
      \midrule
      Base                   & 68.25                                & 85.37                                              & 71.16                                 & 41.40                        & 60.76                        & 47.50                     & 14.78                      & 11.13                        & 82.00                            & 10.00                     & 7.83                              \\
      LoRA                   & 64.88                                & 76.95                                              & 57.12                                 & 34.66                        & 45.82                        & 64.55                     & 21.87                      & 20.79                        & 84.75                            & 73.61                     & 5.05                              \\
      ParallelMLP            & 68.01                                & 83.02                                              & 61.74                                 & 31.84                        & 42.32                        & \textbf{59.33}            & 1.89                       & 12.01                        & \textbf{71.55}$^{**}$            & 38.05                     & 7.22                              \\
      CAL (\textsc{Late8th}) & \textbf{68.25}                       & \textbf{83.02}                                     & \textbf{69.13}$^{*}$                  & \textbf{33.92}               & \textbf{58.60}$^{**}$        & 58.75                     & \textbf{6.66}$^{**}$       & \textbf{14.14}$^{**}$        & 81.90                            & \textbf{25.00}$^{**}$     & \textbf{6.80}                     \\
      \bottomrule
    \end{tabular}}
\end{table}

\paragraph{Results.}
Both frozen-backbone adapters preserve general capability substantially better than LoRA. Against the matched ParallelMLP control, \textsc{late8th} gains 7.4 pp on IFEval and 16.3 pp on SysBench and also improves Glaive FC, IH-Challenge, and MSJ. The advantage is not uniform: SQuAD is nearly unchanged, TensorTrust changes only marginally, and InjecAgent favors ParallelMLP. The dedicated system-prompt route therefore helps most with explicit constraints, multi-turn adherence, and sustained many-shot attacks rather than every form of extraction or prompt injection.

\paragraph{Comparison with the 1.5B study.}
The two adapters are close on most benchmarks at 1.5B, whereas the 8B comparison separates them more clearly on IFEval, SysBench, and MSJ.
\section{Limitations and future work}
\label{sec:discussion}

This study is limited to cross-attention blocks trained on a frozen backbone with one 50,000-example SFT corpus. The observed placement rankings may therefore depend on the corpus, optimization recipe, backbone family, and the decision to freeze all original model weights; they may not generalize to joint training, preference optimization, or broader data distributions. Each configuration was trained once, so the experiments do not measure variation across training runs or random initialization.

Cross-model transfer is also under-studied. We sweep placements only on Qwen2.5-1.5B and train only \textsc{late8th} on Llama-3.1-8B because of the compute budget. The 8B comparison changes the model family, tokenizer, chat template, and hyperparameters together, so it cannot isolate scale or establish that the same placement is optimal at larger sizes. CAL also does not uniformly outperform the unmodified Base model on security benchmarks; the matched ParallelMLP comparison studies the effect of system-prompt routing relative to late-stage parameter addition, rather than establishing CAL as a general security defense.

Future work should study cross-attention placement on other instruction-following and security tasks, including settings in which the privileged information is not a system prompt. Repeating placement sweeps across wider training corpora, model families, scales, and random seeds would test the generality of the current conclusions. Finally, the finding that later placement is often effective and parameter-efficient can inform training procedures designed specifically for stronger security or rule following, including adversarial and preference-based training of the inserted blocks.

\paragraph{Broader impacts.}\label{sec:broader_impacts}
Improved system-instruction routing may reduce some failures in multi-turn and agentic applications, but it is not a hard security boundary and should not replace access control or input sanitization. Publishing attack-success-rate analyses may inform adversaries; however, all benchmarks are existing public resources and this paper introduces no new attack methods. Stronger rule following can also reinforce harmful system instructions.

\section{Conclusion}
\label{sec:conclusion}

We studied how injecting a system-prompt span through cross-attention affects instruction-following and security tasks, with particular attention to where the injected blocks are placed. The 1.5B sweep shows that placement is a major, task-dependent factor: later blocks are generally strong and parameter-efficient. At 8B, the transferred \textsc{late8th} configuration separates more clearly from matched ParallelMLP on IFEval, SysBench, and Many-Shot Jailbreaking. Overall, the results motivate placement-aware use of cross-attention for system-prompt routing and broader validation across tasks, corpora, backbones, and scales.

\begingroup
\small
\bibliographystyle{plainnat}
\bibliography{paper}

@inproceedings{vaswani2017attention,
  author       = {Ashish Vaswani and
                  Noam Shazeer and
                  Niki Parmar and
                  Jakob Uszkoreit and
                  Llion Jones and
                  Aidan N. Gomez and
                  Lukasz Kaiser and
                  Illia Polosukhin},
  editor       = {Isabelle Guyon and
                  Ulrike von Luxburg and
                  Samy Bengio and
                  Hanna M. Wallach and
                  Rob Fergus and
                  S. V. N. Vishwanathan and
                  Roman Garnett},
  title        = {Attention is All you Need},
  booktitle    = {Advances in Neural Information Processing Systems 30: Annual Conference
                  on Neural Information Processing Systems 2017, December 4-9, 2017,
                  Long Beach, CA, {USA}},
  pages        = {5998--6008},
  year         = {2017},
  url          = {https://proceedings.neurips.cc/paper/2017/hash/3f5ee243547dee91fbd053c1c4a845aa-Abstract.html},
  bibsource    = {dblp computer science bibliography, https://dblp.org}
}

@article{wallace2025instruction,
  author       = {Eric Wallace and
                  Kai Xiao and
                  Reimar Leike and
                  Lilian Weng and
                  Johannes Heidecke and
                  Alex Beutel},
  title        = {The Instruction Hierarchy: Training LLMs to Prioritize Privileged
                  Instructions},
  journal      = {CoRR},
  volume       = {abs/2404.13208},
  year         = {2024},
  url          = {https://doi.org/10.48550/arXiv.2404.13208},
  doi          = {10.48550/ARXIV.2404.13208},
  eprinttype   = {arXiv},
  eprint       = {2404.13208},
  bibsource    = {dblp computer science bibliography, https://dblp.org}
}

@article{perez2022ignore,
    author       = {F{\'{a}}bio Perez and
                  Ian Ribeiro},
  title        = {Ignore Previous Prompt: Attack Techniques For Language Models},
  journal      = {CoRR},
  volume       = {abs/2211.09527},
  year         = {2022},
  url          = {https://doi.org/10.48550/arXiv.2211.09527},
  doi          = {10.48550/ARXIV.2211.09527},
  eprinttype   = {arXiv},
  eprint       = {2211.09527},
  bibsource    = {dblp computer science bibliography, https://dblp.org}
}

@inproceedings{greshake2023indirect,
  author       = {Sahar Abdelnabi and
                  Kai Greshake and
                  Shailesh Mishra and
                  Christoph Endres and
                  Thorsten Holz and
                  Mario Fritz},
  editor       = {Maura Pintor and
                  Xinyun Chen and
                  Florian Tram{\`{e}}r},
  title        = {Not What You've Signed Up For: Compromising Real-World LLM-Integrated
                  Applications with Indirect Prompt Injection},
  booktitle    = {Proceedings of the 16th {ACM} Workshop on Artificial Intelligence
                  and Security, AISec 2023, Copenhagen, Denmark, 30 November 2023},
  pages        = {79--90},
  publisher    = {{ACM}},
  year         = {2023},
  url          = {https://doi.org/10.1145/3605764.3623985},
  doi          = {10.1145/3605764.3623985},
  bibsource    = {dblp computer science bibliography, https://dblp.org}
}

@inproceedings{hu2022lora,
    author       = {Edward J. Hu and
                  Yelong Shen and
                  Phillip Wallis and
                  Zeyuan Allen{-}Zhu and
                  Yuanzhi Li and
                  Shean Wang and
                  Lu Wang and
                  Weizhu Chen},
  title        = {LoRA: Low-Rank Adaptation of Large Language Models},
  booktitle    = {The Tenth International Conference on Learning Representations, {ICLR}
                  2022, Virtual Event, April 25-29, 2022},
  publisher    = {OpenReview.net},
  year         = {2022},
  url          = {https://openreview.net/forum?id=nZeVKeeFYf9},
  bibsource    = {dblp computer science bibliography, https://dblp.org}
}

@inproceedings{piet2024secalign,
    author       = {Sizhe Chen and
                  Arman Zharmagambetov and
                  Saeed Mahloujifar and
                  Kamalika Chaudhuri and
                  David A. Wagner and
                  Chuan Guo},
  editor       = {Chun{-}Ying Huang and
                  Jyh{-}Cheng Chen and
                  Shiuh{-}Pyng Shieh and
                  David Lie and
                  V{\'{e}}ronique Cortier},
  title        = {SecAlign: Defending Against Prompt Injection with Preference Optimization},
  booktitle    = {Proceedings of the 2025 {ACM} {SIGSAC} Conference on Computer and
                  Communications Security, {CCS} 2025, Taipei, Taiwan, October 13-17,
                  2025},
  pages        = {2833--2847},
  publisher    = {{ACM}},
  year         = {2025},
  url          = {https://doi.org/10.1145/3719027.3744836},
  doi          = {10.1145/3719027.3744836},
  bibsource    = {dblp computer science bibliography, https://dblp.org}
}

@inproceedings{chen2025struq,
  author       = {Sizhe Chen and
                  Julien Piet and
                  Chawin Sitawarin and
                  David A. Wagner},
  editor       = {Lujo Bauer and
                  Giancarlo Pellegrino},
  title        = {StruQ: Defending Against Prompt Injection with Structured Queries},
  booktitle    = {34th {USENIX} Security Symposium, {USENIX} Security 2025, Seattle,
                  WA, USA, August 13-15, 2025},
  pages        = {2383--2400},
  publisher    = {{USENIX} Association},
  year         = {2025},
  url          = {https://www.usenix.org/conference/usenixsecurity25/presentation/chen-sizhe},
  bibsource    = {dblp computer science bibliography, https://dblp.org}
}

@inproceedings{zhang2023controlnet,
    author       = {Lvmin Zhang and
                  Anyi Rao and
                  Maneesh Agrawala},
  title        = {Adding Conditional Control to Text-to-Image Diffusion Models},
  booktitle    = {{IEEE/CVF} International Conference on Computer Vision, {ICCV} 2023,
                  Paris, France, October 1-6, 2023},
  pages        = {3813--3824},
  publisher    = {{IEEE}},
  year         = {2023},
  url          = {https://doi.org/10.1109/ICCV51070.2023.00355},
  doi          = {10.1109/ICCV51070.2023.00355},
  bibsource    = {dblp computer science bibliography, https://dblp.org}
}

@inproceedings{alayrac2022flamingo,
 author       = {Jean{-}Baptiste Alayrac and
                  Jeff Donahue and
                  Pauline Luc and
                  Antoine Miech and
                  Iain Barr and
                  Yana Hasson and
                  Karel Lenc and
                  Arthur Mensch and
                  Katherine Millican and
                  Malcolm Reynolds and
                  Roman Ring and
                  Eliza Rutherford and
                  Serkan Cabi and
                  Tengda Han and
                  Zhitao Gong and
                  Sina Samangooei and
                  Marianne Monteiro and
                  Jacob L. Menick and
                  Sebastian Borgeaud and
                  Andy Brock and
                  Aida Nematzadeh and
                  Sahand Sharifzadeh and
                  Mikolaj Binkowski and
                  Ricardo Barreira and
                  Oriol Vinyals and
                  Andrew Zisserman and
                  Kar{\'{e}}n Simonyan},
  editor       = {Sanmi Koyejo and
                  S. Mohamed and
                  A. Agarwal and
                  Danielle Belgrave and
                  K. Cho and
                  A. Oh},
  title        = {Flamingo: a Visual Language Model for Few-Shot Learning},
  booktitle    = {Advances in Neural Information Processing Systems 35: Annual Conference
                  on Neural Information Processing Systems 2022, NeurIPS 2022, New Orleans,
                  LA, USA, November 28 - December 9, 2022},
  year         = {2022},
  url          = {http://papers.nips.cc/paper\_files/paper/2022/hash/960a172bc7fbf0177ccccbb411a7d800-Abstract-Conference.html},
  bibsource    = {dblp computer science bibliography, https://dblp.org}
}

@inproceedings{li2023blip2,
  author       = {Junnan Li and
                  Dongxu Li and
                  Silvio Savarese and
                  Steven C. H. Hoi},
  editor       = {Andreas Krause and
                  Emma Brunskill and
                  Kyunghyun Cho and
                  Barbara Engelhardt and
                  Sivan Sabato and
                  Jonathan Scarlett},
  title        = {{BLIP-2:} Bootstrapping Language-Image Pre-training with Frozen Image
                  Encoders and Large Language Models},
  booktitle    = {International Conference on Machine Learning, {ICML} 2023, 23-29 July
                  2023, Honolulu, Hawaii, {USA}},
  series       = {Proceedings of Machine Learning Research},
  pages        = {19730--19742},
  publisher    = {{PMLR}},
  year         = {2023},
  url          = {https://proceedings.mlr.press/v202/li23q.html},
  bibsource    = {dblp computer science bibliography, https://dblp.org}
}

@article{qwen2025,
  title   = {Qwen2.5 Technical Report},
  author  = {An Yang and Baosong Yang and Beichen Zhang and Binyuan Hui and Bo Zheng and
             Bowen Yu and Chengyuan Li and Dayiheng Liu and Fei Huang and Haoran Wei and
             Huan Lin and Jian Yang and Jianhong Tu and Jianwei Zhang and Jianxin Yang and
             Jiaxi Yang and Jingren Zhou and Junyang Lin and Kai Dang and Keming Lu and
             Keqin Bao and Kexin Yang and Le Yu and Mei Li and Mingfeng Xue and Pei Zhang and
             Qin Zhu and Rui Men and Runji Lin and Tianhao Li and Tianyi Tang and Tingyu Xia and
             Xingzhang Ren and Xuancheng Ren and Yang Fan and Yang Su and Yichang Zhang and
             Yu Wan and Yuqiong Liu and Zeyu Cui and Zhenru Zhang and Zihan Qiu},
  journal = {arXiv preprint arXiv:2412.15115},
  year    = {2024},
  doi     = {10.48550/arXiv.2412.15115},
  url     = {https://arxiv.org/abs/2412.15115}
}

@inproceedings{he2021towards,
 author       = {Junxian He and
                  Chunting Zhou and
                  Xuezhe Ma and
                  Taylor Berg{-}Kirkpatrick and
                  Graham Neubig},
  title        = {Towards a Unified View of Parameter-Efficient Transfer Learning},
  booktitle    = {The Tenth International Conference on Learning Representations, {ICLR}
                  2022, Virtual Event, April 25-29, 2022},
  publisher    = {OpenReview.net},
  year         = {2022},
  url          = {https://openreview.net/forum?id=0RDcd5Axok},
  bibsource    = {dblp computer science bibliography, https://dblp.org}
}

@article{cobbe2021gsm8k,
  author       = {Karl Cobbe and
                  Vineet Kosaraju and
                  Mohammad Bavarian and
                  Mark Chen and
                  Heewoo Jun and
                  Lukasz Kaiser and
                  Matthias Plappert and
                  Jerry Tworek and
                  Jacob Hilton and
                  Reiichiro Nakano and
                  Christopher Hesse and
                  John Schulman},
  title        = {Training Verifiers to Solve Math Word Problems},
  journal      = {CoRR},
  volume       = {abs/2110.14168},
  year         = {2021},
  url          = {https://arxiv.org/abs/2110.14168},
  eprinttype   = {arXiv},
  eprint       = {2110.14168},
  bibsource    = {dblp computer science bibliography, https://dblp.org}
}

@inproceedings{hendrycks2021mmlu,
  author       = {Dan Hendrycks and
                  Collin Burns and
                  Steven Basart and
                  Andy Zou and
                  Mantas Mazeika and
                  Dawn Song and
                  Jacob Steinhardt},
  title        = {Measuring Massive Multitask Language Understanding},
  booktitle    = {9th International Conference on Learning Representations, {ICLR} 2021,
                  Virtual Event, Austria, May 3-7, 2021},
  publisher    = {OpenReview.net},
  year         = {2021},
  url          = {https://openreview.net/forum?id=d7KBjmI3GmQ},
  bibsource    = {dblp computer science bibliography, https://dblp.org}
}

@article{zhou2023ifeval,
  author       = {Jeffrey Zhou and
                  Tianjian Lu and
                  Swaroop Mishra and
                  Siddhartha Brahma and
                  Sujoy Basu and
                  Yi Luan and
                  Denny Zhou and
                  Le Hou},
  title        = {Instruction-Following Evaluation for Large Language Models},
  journal      = {CoRR},
  volume       = {abs/2311.07911},
  year         = {2023},
  url          = {https://doi.org/10.48550/arXiv.2311.07911},
  doi          = {10.48550/ARXIV.2311.07911},
  eprinttype   = {arXiv},
  eprint       = {2311.07911},
  bibsource    = {dblp computer science bibliography, https://dblp.org}
}

@article{sysbench,
  author       = {Yanzhao Qin and
                  Tao Zhang and
                  Tao Zhang and
                  Yanjun Shen and
                  Wenjing Luo and
                  Haoze Sun and
                  Yan Zhang and
                  Yujing Qiao and
                  Weipeng Chen and
                  Zenan Zhou and
                  Wentao Zhang and
                  Bin Cui},
  title        = {SysBench: Can Large Language Models Follow System Messages?},
  journal      = {CoRR},
  volume       = {abs/2408.10943},
  year         = {2024},
  url          = {https://doi.org/10.48550/arXiv.2408.10943},
  doi          = {10.48550/ARXIV.2408.10943},
  eprinttype   = {arXiv},
  eprint       = {2408.10943},
  bibsource    = {dblp computer science bibliography, https://dblp.org}
}

@inproceedings{rajpurkar2018know,
  author       = {Pranav Rajpurkar and
                  Robin Jia and
                  Percy Liang},
  editor       = {Iryna Gurevych and
                  Yusuke Miyao},
  title        = {Know What You Don't Know: Unanswerable Questions for SQuAD},
  booktitle    = {Proceedings of the 56th Annual Meeting of the Association for Computational
                  Linguistics, {ACL} 2018, Melbourne, Australia, July 15-20, 2018, Volume
                  2: Short Papers},
  pages        = {784--789},
  publisher    = {Association for Computational Linguistics},
  year         = {2018},
  url          = {https://aclanthology.org/P18-2124/},
  doi          = {10.18653/V1/P18-2124},
  bibsource    = {dblp computer science bibliography, https://dblp.org}
}

@misc{glaive2023functioncalling,
  title        = {Glaive Function Calling V2},
  author       = {{Glaive AI}},
  year         = {2023},
  howpublished = {\url{https://huggingface.co/datasets/glaiveai/glaive-function-calling-v2}},
  note         = {Hugging Face dataset, accessed 2026-08-14}
}

@article{ihchallenge,
  author       = {Chuan Guo and
                  Juan Felipe Ceron Uribe and
                  Sicheng Zhu and
                  Christopher A. Choquette{-}Choo and
                  Steph Lin and
                  Nikhil Kandpal and
                  Milad Nasr and
                  Rai (Michael Pokorny) and
                  Sam Toyer and
                  Miles Wang and
                  Yaodong Yu and
                  Alex Beutel and
                  Kai Xiao},
  title        = {IH-Challenge: {A} Training Dataset to Improve Instruction Hierarchy
                  on Frontier LLMs},
  journal      = {CoRR},
  volume       = {abs/2603.10521},
  year         = {2026},
  url          = {https://doi.org/10.48550/arXiv.2603.10521},
  doi          = {10.48550/ARXIV.2603.10521},
  eprinttype   = {arXiv},
  eprint       = {2603.10521},
  bibsource    = {dblp computer science bibliography, https://dblp.org}
}

@inproceedings{zhan2024injecagent,
  author       = {Qiusi Zhan and
                  Zhixiang Liang and
                  Zifan Ying and
                  Daniel Kang},
  editor       = {Lun{-}Wei Ku and
                  Andre Martins and
                  Vivek Srikumar},
  title        = {InjecAgent: Benchmarking Indirect Prompt Injections in Tool-Integrated
                  Large Language Model Agents},
  booktitle    = {Findings of the Association for Computational Linguistics, {ACL} 2024,
                  Bangkok, Thailand and virtual meeting, August 11-16, 2024},
  series       = {Findings of {ACL}},
  pages        = {10471--10506},
  publisher    = {Association for Computational Linguistics},
  year         = {2024},
  url          = {https://doi.org/10.18653/v1/2024.findings-acl.624},
  doi          = {10.18653/V1/2024.FINDINGS-ACL.624},
  bibsource    = {dblp computer science bibliography, https://dblp.org}
}

@inproceedings{anil2024msj,
  author       = {Cem Anil and
                  Esin Durmus and
                  Nina Panickssery and
                  Mrinank Sharma and
                  Joe Benton and
                  Sandipan Kundu and
                  Joshua Batson and
                  Meg Tong and
                  Jesse Mu and
                  Daniel Ford and
                  Francesco Mosconi and
                  Rajashree Agrawal and
                  Rylan Schaeffer and
                  Naomi Bashkansky and
                  Samuel Svenningsen and
                  Mike Lambert and
                  Ansh Radhakrishnan and
                  Carson Denison and
                  Evan J. Hubinger and
                  Yuntao Bai and
                  Trenton Bricken and
                  Timothy Maxwell and
                  Nicholas Schiefer and
                  James Sully and
                  Alex Tamkin and
                  Tamera Lanham and
                  Karina Nguyen and
                  Tomasz Korbak and
                  Jared Kaplan and
                  Deep Ganguli and
                  Samuel R. Bowman and
                  Ethan Perez and
                  Roger Baker Grosse and
                  David Duvenaud},
  editor       = {Amir Globerson and
                  Lester Mackey and
                  Danielle Belgrave and
                  Angela Fan and
                  Ulrich Paquet and
                  Jakub M. Tomczak and
                  Cheng Zhang},
  title        = {Many-shot Jailbreaking},
  booktitle    = {Advances in Neural Information Processing Systems 37: Annual Conference
                  on Neural Information Processing Systems 2024, NeurIPS 2024, Vancouver,
                  BC, Canada, December 10 - 15, 2024},
  volume       = {37},
  year         = {2024},
  doi          = {10.52202/079017-4121},
  url          = {https://proceedings.neurips.cc/paper_files/paper/2024/hash/ea456e232efb72d261715e33ce25f208-Abstract-Conference.html},
  bibsource    = {dblp computer science bibliography, https://dblp.org}
}

@inproceedings{tensortrust,
  author       = {Sam Toyer and
                  Olivia Watkins and
                  Ethan Adrian Mendes and
                  Justin Svegliato and
                  Luke Bailey and
                  Tiffany Wang and
                  Isaac Ong and
                  Karim Elmaaroufi and
                  Pieter Abbeel and
                  Trevor Darrell and
                  Alan Ritter and
                  Stuart Russell},
  title        = {Tensor Trust: Interpretable Prompt Injection Attacks from an Online
                  Game},
  booktitle    = {The Twelfth International Conference on Learning Representations,
                  {ICLR} 2024, Vienna, Austria, May 7-11, 2024},
  publisher    = {OpenReview.net},
  year         = {2024},
  url          = {https://openreview.net/forum?id=fsW7wJGLBd},
  bibsource    = {dblp computer science bibliography, https://dblp.org}
}

@misc{kamradt2023needle,
  author       = {Greg Kamradt},
  title        = {Needle in a Haystack: Pressure Testing {LLM}s},
  year         = {2023},
  howpublished = {GitHub repository},
  url          = {https://github.com/gkamradt/needle-in-a-haystack},
  note         = {Accessed 2026-08-14}
}

@misc{graham2026essays,
  author       = {Paul Graham},
  title        = {Essays},
  year         = {n.d.},
  howpublished = {Personal website},
  url          = {https://www.paulgraham.com/articles.html},
  note         = {Accessed 2026-08-14}
}

@misc{msjdataset,
  author       = {{KutalVolkan}},
  title        = {Many-Shot Jailbreaking Dataset},
  year         = {2024},
  howpublished = {GitHub repository},
  url          = {https://github.com/KutalVolkan/many-shot-jailbreaking-dataset},
  note         = {Commit 5eac855d63bf2261d997dafea0eb8bfb06f8f165; accessed 2026-08-14; no license file stated}
}

@article{zou2023representation,
author       = {Andy Zou and
                  Long Phan and
                  Sarah Li Chen and
                  James Campbell and
                  Phillip Guo and
                  Richard Ren and
                  Alexander Pan and
                  Xuwang Yin and
                  Mantas Mazeika and
                  Ann{-}Kathrin Dombrowski and
                  Shashwat Goel and
                  Nathaniel Li and
                  Michael J. Byun and
                  Zifan Wang and
                  Alex Mallen and
                  Steven Basart and
                  Sanmi Koyejo and
                  Dawn Song and
                  Matt Fredrikson and
                  J. Zico Kolter and
                  Dan Hendrycks},
  title        = {Representation Engineering: {A} Top-Down Approach to {AI} Transparency},
  journal      = {CoRR},
  volume       = {abs/2310.01405},
  year         = {2023},
  url          = {https://doi.org/10.48550/arXiv.2310.01405},
  doi          = {10.48550/ARXIV.2310.01405},
  eprinttype   = {arXiv},
  eprint       = {2310.01405},
  bibsource    = {dblp computer science bibliography, https://dblp.org}
}

@inproceedings{geva2022transformer,
  author       = {Mor Geva and
                  Avi Caciularu and
                  Kevin Ro Wang and
                  Yoav Goldberg},
  editor       = {Yoav Goldberg and
                  Zornitsa Kozareva and
                  Yue Zhang},
  title        = {Transformer Feed-Forward Layers Build Predictions by Promoting Concepts
                  in the Vocabulary Space},
  booktitle    = {Proceedings of the 2022 Conference on Empirical Methods in Natural
                  Language Processing, {EMNLP} 2022, Abu Dhabi, United Arab Emirates,
                  December 7-11, 2022},
  pages        = {30--45},
  publisher    = {Association for Computational Linguistics},
  year         = {2022},
  url          = {https://doi.org/10.18653/v1/2022.emnlp-main.3},
  doi          = {10.18653/V1/2022.EMNLP-MAIN.3},
  bibsource    = {dblp computer science bibliography, https://dblp.org}
}

@misc{llama3technicalreport,
  title={The Llama 3 Herd of Models}, 
  author={Meta},
  year={2024},
  eprint={2407.21783},
  archivePrefix={arXiv},
  primaryClass={cs.LG},
  url={https://arxiv.org/abs/2407.21783}
}

@misc{OpenHermes2.5,
  title = {OpenHermes 2.5: An Open Dataset of Synthetic Data for Generalist LLM Assistants},
  author = {Teknium},
  year = {2023},
  publisher = {HuggingFace},
  url = {https://huggingface.co/datasets/teknium/OpenHermes-2.5},
  note = {Hugging Face dataset, accessed 2026-08-14}
}

@misc{Airoboros,
  title = {Airoboros: A diverse, synthetically generated dataset for instruction tuning},
  author = {Durbin, Jon},
  year = {2023},
  publisher = {HuggingFace},
  url = {https://huggingface.co/datasets/jondurbin/airoboros-3.2},
  note = {Hugging Face dataset, accessed 2026-08-14}
}

@article{ji2024beavertails,
  title   = {BeaverTails: Towards Improved Safety Alignment of {LLM} via a Human-Preference Dataset},
  author  = {Jiaming Ji and Mickel Liu and Josef Dai and Xuehai Pan and Chi Zhang and Ce Bian and Boyuan Chen and Ruiyang Sun and Yizhou Wang and Yaodong Yang},
  journal = {Advances in Neural Information Processing Systems},
  volume  = {36},
  year    = {2023},
  doi     = {10.52202/075280-1072},
  url     = {https://proceedings.neurips.cc/paper_files/paper/2023/hash/4dbb61cb68671edc4ca3712d70083b9f-Abstract-Datasets_and_Benchmarks.html}
}

@inproceedings{rajbhandari2020zero,
  author       = {Samyam Rajbhandari and
                  Jeff Rasley and
                  Olatunji Ruwase and
                  Yuxiong He},
  editor       = {Christine Cuicchi and
                  Irene Qualters and
                  William T. Kramer},
  title        = {ZeRO: memory optimizations toward training trillion parameter models},
  booktitle    = {Proceedings of the International Conference for High Performance Computing,
                  Networking, Storage and Analysis, {SC} 2020, Virtual Event / Atlanta,
                  Georgia, USA, November 9-19, 2020},
  pages        = {20},
  publisher    = {{IEEE/ACM}},
  year         = {2020},
  url          = {https://doi.org/10.1109/SC41405.2020.00024},
  doi          = {10.1109/SC41405.2020.00024},
  bibsource    = {dblp computer science bibliography, https://dblp.org}
}

@inproceedings{loshchilov2019adamw,
  author       = {Ilya Loshchilov and
                  Frank Hutter},
  title        = {Decoupled Weight Decay Regularization},
  booktitle    = {7th International Conference on Learning Representations, {ICLR} 2019,
                  New Orleans, LA, USA, May 6-9, 2019},
  publisher    = {OpenReview.net},
  year         = {2019},
  url          = {https://openreview.net/forum?id=Bkg6RiCqY7},
  bibsource    = {dblp computer science bibliography, https://dblp.org}
}

@inproceedings{zhang2019rmsnorm,
  title     = {Root Mean Square Layer Normalization},
  author    = {Zhang, Biao and Sennrich, Rico},
  booktitle = {Advances in Neural Information Processing Systems},
  volume    = {32},
  year      = {2019},
  note      = {arXiv:1910.07467}
}

@article{dao2023flashattn2,
  title   = {{FlashAttention-2}: Faster Attention with Better Parallelism and Work Partitioning},
  author  = {Dao, Tri},
  journal = {arXiv preprint arXiv:2307.08691},
  year    = {2023},
  doi     = {10.48550/arXiv.2307.08691},
  url     = {https://arxiv.org/abs/2307.08691}
}

@article{shazeer2020swiglu,
  title   = {{GLU} Variants Improve Transformer},
  author  = {Shazeer, Noam},
  journal = {arXiv preprint arXiv:2002.05202},
  year    = {2020}
}

@article{raffel2019t5,
  title   = {Exploring the Limits of Transfer Learning with a Unified Text-to-Text Transformer},
  author  = {Raffel, Colin and Shazeer, Noam and Roberts, Adam and Lee, Katherine and Narang, Sharan and Matena, Michael and Zhou, Yanqi and Li, Wei and Liu, Peter J.},
  journal = {Journal of Machine Learning Research},
  volume  = {21},
  number  = {140},
  pages   = {1--67},
  year    = {2020},
  note    = {arXiv:1910.10683}
}

@inproceedings{borgeaud2022retro,
  title     = {Improving Language Models by Retrieving from Trillions of Tokens},
  author    = {Borgeaud, Sebastian and Mensch, Arthur and Hoffmann, Jordan and Cai, Trevor and Rutherford, Eliza and Millican, Katie and van den Driessche, George and Lespiau, Jean-Baptiste and Damoc, Bogdan and Clark, Aidan and others},
  booktitle = {Proceedings of the 39th International Conference on Machine Learning},
  pages     = {2206--2240},
  year      = {2022}
}

@article{kariyappa2025air,
  title   = {Stronger Enforcement of Instruction Hierarchy via Augmented Intermediate Representations},
  author  = {Kariyappa, Sanjay and Suh, G. Edward},
  journal = {arXiv preprint arXiv:2505.18907},
  year    = {2025},
  doi     = {10.48550/arXiv.2505.18907}
}

@inproceedings{zverev2026aside,
  title     = {{ASIDE}: Architectural Separation of Instructions and Data in Language Models},
  author    = {Zverev, Egor and Kortukov, Evgenii and Panfilov, Alexander and Volkova, Alexandra and Tabesh, Soroush and Lapuschkin, Sebastian and Samek, Wojciech and Lampert, Christoph H.},
  booktitle = {International Conference on Learning Representations},
  year      = {2026}
}
\endgroup

\appendix

\section{Implementation details}
\label{app:impl}

The following subsections detail measured inference cost, CAL placement configurations, training hyperparameters, dataset composition, system-prompt span detection, batched processing, and parameter accounting.

\subsection{Measured inference cost}
\label{app:inference_cost}

On Qwen2.5-1.5B at batch size one, five \textsc{late8th} CAL blocks add 188.76M parameters (12.23\%) and 360 MB of bf16 weights. Across 512--4096-token prompts, median prefill overhead is 18.2--19.9\%, time-to-first-token overhead is 17.9--19.9\%, and decode overhead is 12.0--13.2\%:
\begin{center}
  \small
  \begin{tabular}{lrrrrr}
    \toprule
                 & Params     & Weights   & Prefill           & TTFT              & Decode            \\
    \midrule
    CAL $-$ Base & $+12.23\%$ & $+360$ MB & $+18.2$--$19.9\%$ & $+17.9$--$19.9\%$ & $+12.0$--$13.2\%$ \\
    \bottomrule
  \end{tabular}
\end{center}
Measurements use bf16 FlashAttention-2~\citep{dao2023flashattn2}, greedy decoding, 256 generated tokens, 20 interleaved runs, and a system span occupying half of each prompt on an RTX 4080 Laptop GPU. The nearly stable percentage overhead across an $8\times$ prompt-length range is consistent with caching CAL keys/values, while the added layers impose a real fixed cost. Supplying span bounds directly also avoids a Python token-scan overhead of 1.7--13.8 ms.

\subsection{CAL placement configurations}
\label{app:config}
Table~\ref{tab:app_configs} lists the ten specific CAL placement strategies evaluated across the 28-layer backbone to map the spatial locality of system-prompt constraints.

\begin{table}[ht]
  \caption{Detailed CAL placement configurations evaluated on the 28-layer Qwen2.5-1.5B-Instruct backbone.}
  \label{tab:app_configs}
  \centering
  \begin{tabular}{llc}
    \toprule
    \textbf{Configuration} & \textbf{Placement Strategy}                                         & \textbf{Blocks ($|\mathcal{P}|$)} \\
    \midrule
    \textsc{all}           & After every layer                                                   & 28                                \\
    \textsc{every2}        & After every other layer                                             & 14                                \\
    \textsc{earlythird}    & Contiguous 11-block window at layers 1--11                          & 11                                \\
    \textsc{midthird}      & Contiguous 11-block window at layers 9--19                          & 11                                \\
    \textsc{latethird}     & Contiguous 11-block window at layers 18--28                         & 11                                \\
    \textsc{latehalf}      & Distributed across the last $\nicefrac{1}{2}$ of layers, contiguous & 15                                \\
    \textsc{latehalf2}     & Interleaved throughout the last $\nicefrac{1}{2}$ of layers         & 8                                 \\
    \textsc{latequarter}   & Distributed across the last $\nicefrac{1}{4}$ of layers             & 8                                 \\
    \textsc{late8th}       & Distributed across the last $\nicefrac{1}{8}$ of layers             & 5                                 \\
    \textsc{last}          & A single block after the final layer                                & 1                                 \\
    \bottomrule
  \end{tabular}
\end{table}

\subsection{Training hyperparameters and hardware}
\label{app:hparams}

Table~\ref{tab:hparams} lists the full set of hyperparameters used to train the CAL adapter layers for both experimental phases. All runs use a cosine learning-rate schedule with a linear warmup phase, fused AdamW~\citep{loshchilov2019adamw}, and bfloat16 mixed precision. Training is distributed via DeepSpeed ZeRO Stage~2~\citep{rajbhandari2020zero} with CPU optimizer-state offload, which reduces peak VRAM by approximately 50\% compared to a standard data-parallel setup. Gradient checkpointing was disabled in the reported runs. Freezing the backbone prevents weight-gradient storage, but activation gradients still propagate from each inserted CAL block through later decoder layers; we therefore make no training-speed claim from freezing alone. Each configuration was trained once, so the results do not measure variability across initialization seeds or training-data orderings. All model training, ablation runs, and benchmark evaluations were conducted using NVIDIA A100 80GB hardware, requiring approximately 20 total compute days across the project. The backbone models are used under their respective licenses: Qwen2.5-1.5B-Instruct is released under the Apache~2.0 license; Llama-3.1-8B-Instruct is released under the Llama~3.1 Community License.

\begin{table}[ht]
  \caption{Full training hyperparameter settings for each backbone. The conservative Llama~3.1 settings (lower LR, longer warmup, tighter gradient clipping) mitigate catastrophic forgetting in the larger model.}
  \label{tab:hparams}
  \centering
  \begin{tabular}{lcc}
    \toprule
    \textbf{Setting}             & \textbf{Qwen2.5-1.5B} & \textbf{Llama-3.1-8B} \\
    \midrule
    Learning rate                & $2\!\times\!10^{-4}$  & $5\!\times\!10^{-5}$  \\
    LR schedule                  & cosine                & cosine                \\
    Warmup ratio                 & 0.05                  & 0.10                  \\
    Weight decay                 & 0.01                  & 0.01                  \\
    Optimizer                    & AdamW (fused)         & AdamW (fused)         \\
    Per-device batch size        & 4                     & 4                     \\
    Gradient accumulation steps  & 32                    & 32                    \\
    Effective batch size         & 128                   & 128                   \\
    Max gradient norm            & 1.0                   & 0.5                   \\
    Training epochs              & 2                     & 2                     \\
    Independent training runs    & 1                     & 1                     \\
    Max sequence length (tokens) & 2{,}048               & 2{,}048               \\
    Precision                    & bfloat16              & bfloat16              \\
    Gradient checkpointing       & disabled              & disabled              \\
    \bottomrule
  \end{tabular}
\end{table}

\subsection{System prompt span detection}
\label{app:detection}

CAL blocks require the token-level span $[s, e)$ of the system prompt at runtime. The following procedure is applied at every prefill step, supporting the ChatML format used by Qwen~\citep{qwen2025} and the header format used by Llama~3~\citep{llama3technicalreport}. The returned interval is left-closed, right-open (Python-slice convention): \texttt{hidden\_states[:, s:e, :]} spans from the opening delimiter through the closing delimiter inclusive.

\begin{enumerate}
  \item Decode each token $t_i$ in the sequence to its string representation.
  \item \textbf{Start detection:} if \texttt{"system"} appears in the current decoded string \emph{and} the previous token string is \texttt{<|im\_start|>} (ChatML) or \texttt{<|start\_header\_id|>} (Llama-3 header), record $s = i - 1$ (the inclusive index of the opening delimiter).
  \item \textbf{End detection:} once inside the system-prompt span, if the current token string contains \texttt{<|im\_end|>} or \texttt{<|eot\_id|>}, record $e = i + 1$ (one past the closing delimiter) and return $(s,\, e)$.
  \item If no system prompt is found, return $(0,\, 0)$; the CAL cross-attention is a no-op for that sample and its output is zero-masked.
\end{enumerate}

For batched inference, \texttt{get\_batch\_sys\_bounds} applies this procedure to each row and returns a \texttt{LongTensor} of shape $(B, 2)$. Rows with no system prompt receive $[0, 0]$. Span boundaries are cached after the prefill step and reused for all subsequent autoregressive decoding steps without recomputation.

\subsection{Batched processing and generation}
\label{app:batching}

For heterogeneous batches, system-prompt keys and values are zero-padded to the maximum span length $\ell_{\max}$ in the batch. A per-sample mask combines the causal rule $j\leq i-s_b$ with a padding mask that assigns $-\infty$ to slots beyond sample $b$'s true span length $\ell_{\mathrm{sys},b}$. Causality preserves the decoder semantics of the system-span representations: a system token cannot depend on a later system token. Once the complete system span precedes a query, user-context and generated tokens can attend to all of it. Thus, the mask prevents future-token leakage while constructing the privileged memory; it does not hide the completed system prompt from later tokens. During autoregressive generation, system-span bounds and CAL keys and values are computed during prefill and cached for subsequent decoding steps.

\subsection{Baseline parameter accounting}
\label{app:lora}

The LoRA~\citep{hu2022lora} baselines target the trainable-parameter budget of the five-block \textsc{late8th} CAL configuration. A CAL block contains four $H\times H$ attention projections, a SwiGLU sublayer of width $4H$, and two RMSNorm vectors, for $16H^2+2H$ trainable parameters per block. Given the base model's hidden dimension $H$, GQA KV dimension $d_\mathrm{kv}$, and backbone feed-forward intermediate size $I$, the equal-rank LoRA solution across all $n$ layers is:
\begin{equation}
  r = \left\lfloor \frac{P_\mathrm{target}}{n \cdot \bigl(2(H + d_\mathrm{kv}) + 4H + 3(H + I)\bigr)} \right\rceil
  \label{eq:lora_rank}
\end{equation}
where $P_\mathrm{target}$ is the trainable parameter budget of the corresponding CAL configuration and $\lfloor\cdot\rceil$ denotes rounding to the nearest integer. When $r > d_\mathrm{kv}$ (the GQA KV-projection ceiling), \texttt{k\_proj} and \texttt{v\_proj} are capped at rank $d_\mathrm{kv}$ and the residual budget is redistributed to the remaining modules. All LoRA adapters target the seven linear projections in each transformer block (\texttt{q\_proj}, \texttt{k\_proj}, \texttt{v\_proj}, \texttt{o\_proj}, \texttt{gate\_proj}, \texttt{up\_proj}, \texttt{down\_proj}), use $\alpha = 2r$, and apply zero dropout.

Equation~\eqref{eq:lora_rank} gives $r=164$ ($\alpha=328$) for Qwen2.5-1.5B and $r=512$ ($\alpha=1024$) for Llama-3.1-8B. Table~\ref{tab:param_counts} reports the resulting budgets. Small differences arise from integer rank and width rounding. ParallelMLP uses five adapters at the same locations as \textsc{late8th}; its intermediate width is chosen by the same target-budget calculation. These counts are computed from the dimensions and target modules instantiated by the training code.

\begin{table}[t]
  \caption{Trainable parameter budgets for the \textsc{late8th} comparison. Counts exclude the frozen backbone.}
  \label{tab:param_counts}
  \centering
  \small
  \begin{tabular}{lrrr}
    \toprule
    \textbf{Backbone} & \textbf{CAL} & \textbf{LoRA} & \textbf{ParallelMLP} \\
    \midrule
    Qwen2.5-1.5B & 188,759,040 & 189,263,872 & 188,751,360 \\
    Llama-3.1-8B & 1,342,218,240 & 1,342,177,280 & 1,342,197,760 \\
    \bottomrule
  \end{tabular}
\end{table}

\subsection{Dataset composition}
\label{app:data}

Table~\ref{tab:data} lists the exact sample counts per source after filtering. Sequences exceeding the 2{,}048-token limit are truncated; no padding is applied during tokenization. All sources are shuffled with seed 42 before selection.

\begin{table}[ht]
  \caption{SFT corpus composition (40\% OpenHermes / 40\% Airoboros / 20\% PKU-SafeRLHF).}
  \label{tab:data}
  \centering
  \small
  \begin{tabular}{lrp{4.2cm}l}
    \toprule
    \textbf{Source}                        & \textbf{Samples}  & \textbf{Filter / selection rule} & \textbf{License} \\
    \midrule
    OpenHermes-2.5~\citep{OpenHermes2.5}   & 20{,}000          &
    Rows grouped by unique system-prompt text; examples drawn by round-robin across groups
    to maximize system-prompt diversity. Rows lacking a system turn receive the default
    prompt \emph{``You are a helpful AI assistant.''}
                                           & Not stated                                                              \\[4pt]
    Airoboros-3.2~\citep{Airoboros}        & 20{,}000          &
    Rows must include all three conversation roles (\texttt{system}, \texttt{human},
    \texttt{gpt}); qualifying rows are shuffled and the first 20{,}000 retained.
                                           & CC BY 4.0                                                               \\[4pt]
    PKU-SafeRLHF~\citep{ji2024beavertails} & 10{,}000          &
    Only rows where the higher-quality (chosen) response satisfies \texttt{is\_safe\,=\,True}
    and \texttt{severity\_level\,=\,0} (both fields refer to the chosen response) are retained. Per-example system prompt randomly drawn from
    a pool of 20 distinct safety phrasings to prevent the CAL layer from memorizing
    a fixed phrasing.
                                           & CC BY-NC 4.0                                                            \\
    \midrule
    \textbf{Total}                         & \textbf{50{,}000} &                                  &                  \\
    \bottomrule
  \end{tabular}
\end{table}

\subsection{Long-IFEval construction}
\label{app:longifeval}

Long-IFEval uses all 541 prompts from the public IFEval split~\citep{zhou2023ifeval}. For each example, the original IFEval prompt is placed verbatim in the system-message field. The user message consists of a deterministic essay-text prefix, the separator \texttt{---}, and the fixed request \emph{``Solve the task described in the system prompt.''} The essay source list contains 218 URLs: 169 pages from Paul Graham's essay index~\citep{graham2026essays} and 49 plain-text mirrors in the Needle-in-a-Haystack repository~\citep{kamradt2023needle}. HTML pages are fetched once, script and style blocks and remaining tags are removed, common HTML entities are decoded, and the resulting plain text is cached together with its source URL. Successfully retrieved essays are concatenated in URL-list order with blank-line separators; no sampling or shuffling is used.

For a target budget $L\in\{4096,8192,16384,32768\}$, the haystack is truncated at a tokenizer boundary after reserving the tokens occupied by the IFEval system prompt and fixed user request, 1,024 tokens for generation, and a 64-token allowance for chat-template control tokens. Generation is greedy (\texttt{do\_sample=False}) with at most 1,024 new tokens. Responses are evaluated with the original IFEval verifier, and the paper reports prompt-level strict accuracy. Thus the length label denotes the target total context budget, including the reserved generation budget, rather than an assertion that every serialized input contains exactly $L$ tokens.

\subsection{Benchmark assets and licenses}
\label{app:benchmark_licenses}

Table~\ref{tab:benchmark_licenses} records the license stated by each benchmark's official distribution, or explicitly notes when no license was stated. The experiments use these assets for research evaluation and this submission does not redistribute their data.

SysBench and MSJ are the only evaluations scored with an LLM judge. Both use the DeepSeek API model \texttt{deepseek-chat}: SysBench uses the benchmark's evaluation prompts with temperature 1.0, while MSJ uses a binary \textsc{complied}/\textsc{refused} safety prompt with temperature 0.

\begin{table}[ht]
  \caption{Evaluation assets and license information stated by their distributions. ``Not stated'' means that no license declaration was located in the cited distribution; it does not imply that the asset is unrestricted.}
  \label{tab:benchmark_licenses}
  \centering
  \small
  \begin{tabular}{p{2.7cm}p{6.8cm}p{2.8cm}}
    \toprule
    \textbf{Asset}                              & \textbf{Distribution / provenance}                                                                                                                                                         & \textbf{Stated license}          \\
    \midrule
    MMLU~\citep{hendrycks2021mmlu}              & Official \href{https://github.com/hendrycks/test}{\texttt{hendrycks/test}} repository                                                                                                      & MIT                              \\
    GSM8K~\citep{cobbe2021gsm8k}                & Official \href{https://github.com/openai/grade-school-math}{\texttt{openai/grade-school-math}} repository                                                                                  & MIT                              \\
    IFEval~\citep{zhou2023ifeval}               & \href{https://github.com/google-research/google-research/tree/master/instruction_following_eval}{\texttt{google-research}} distribution                                                    & Apache 2.0                       \\
    Long-IFEval                                 & IFEval plus the sources described in Appendix~\ref{app:longifeval}; the Needle-in-a-Haystack software is MIT-licensed, but no public content license was located for the underlying essays & Mixed; essay text not stated     \\
    SysBench~\citep{sysbench}                   & Official \href{https://github.com/PKU-Baichuan-MLSystemLab/SysBench}{\texttt{PKU-Baichuan-MLSystemLab/SysBench}} repository                                                                & Not stated                       \\
    SQuAD 2.0~\citep{rajpurkar2018know}         & Official \href{https://rajpurkar.github.io/SQuAD-explorer/}{Stanford SQuAD} distribution                                                                                                   & CC BY-SA 4.0                     \\
    Glaive FC~\citep{glaive2023functioncalling} & \href{https://huggingface.co/datasets/glaiveai/glaive-function-calling-v2}{\texttt{glaiveai/glaive-function-calling-v2}}                                                                   & Apache 2.0                       \\
    IH-Challenge~\citep{ihchallenge}            & \href{https://huggingface.co/datasets/openai/ih-challenge}{\texttt{openai/ih-challenge}}                                                                                                   & Apache 2.0                       \\
    InjecAgent~\citep{zhan2024injecagent}       & Official \href{https://github.com/uiuc-kang-lab/InjecAgent}{\texttt{uiuc-kang-lab/InjecAgent}} repository                                                                                  & MIT                              \\
    MSJ~\citep{anil2024msj,msjdataset}          & \href{https://github.com/KutalVolkan/many-shot-jailbreaking-dataset}{MSJ dataset mirror}, derived from HarmBench                                                                           & Mirror not stated; HarmBench MIT \\
    TensorTrust~\citep{tensortrust}             & Official \href{https://github.com/HumanCompatibleAI/tensor-trust}{\texttt{HumanCompatibleAI/tensor-trust}} repository                                                                      & BSD 2-Clause                     \\
    \bottomrule
  \end{tabular}
\end{table}

\end{document}